\documentclass[runningheads]{llncs}
\usepackage{graphicx}
\usepackage{amsmath}
\usepackage{amsfonts}
\usepackage{kotex}
\usepackage{comment}
\usepackage{subfig}

\usepackage{xcolor}

\usepackage{subfig}
\usepackage{booktabs}
\usepackage{multirow}
\usepackage{tablefootnote}
\usepackage{booktabs}
\usepackage{blindtext}
\usepackage{float}

\begin{document}
%
\title{Multiple Areal Feature Aware Transportation Demand Prediction}
%
%

\author{Sumin Han\inst{1} \and Jisun An\inst{2} \and Youngjun Park\inst{1} \and Suji Kim\inst{3} \and Kitae Jang\inst{3} \and Dongman Lee\inst{1}}

\institute{%
  School of Computing, Korea Advanced Institute of Science and Technology (KAIST)\\
  \email{\{hsm6911,youngjourpark,dlee\}@kaist.ac.kr}
\and
Luddy School of Informatics, Computing, and, Engineering
  Indiana University Bloomington \\
  \email{jisunan@iu.edu}
\and
  The Cho Chun Shik Graduate School of Mobility, Korea Advanced Institute of Science and Technology (KAIST)\\
  \email{\{sujikim,kitaejang\}@kaist.ac.kr}
}

\authorrunning{Sumin Han et al.}
%
\maketitle              
\begin{abstract}
A reliable short-term transportation demand prediction supports the authorities in improving the capability of systems by optimizing schedules, adjusting fleet sizes, and generating new transit networks.
A handful of research efforts incorporate one or a few areal features while learning spatio-temporal correlation, to capture similar demand patterns between similar areas.
However, urban characteristics are polymorphic, and they need to be understood by multiple areal features such as land use, sociodemographics, and place-of-interest (POI) distribution.
In this paper, we propose a novel spatio-temporal multi-feature-aware graph convolutional recurrent network (ST-MFGCRN) that fuses multiple areal features during spatio-temproal understanding.
Inside ST-MFGCRN, we devise sentinel attention to calculate the areal similarity matrix by allowing each area to take partial attention if the feature is not useful.
We evaluate the proposed model on two real-world transportation datasets, one with our constructed BusDJ dataset and one with benchmark TaxiBJ.
Results show that our model outperforms the state-of-the-art baselines up to 7\% on BusDJ and 8\% on TaxiBJ dataset.

\end{abstract}

\section{Introduction}

\begin{figure}[t]
    \centering
    \includegraphics[width=1\columnwidth,height=5cm]{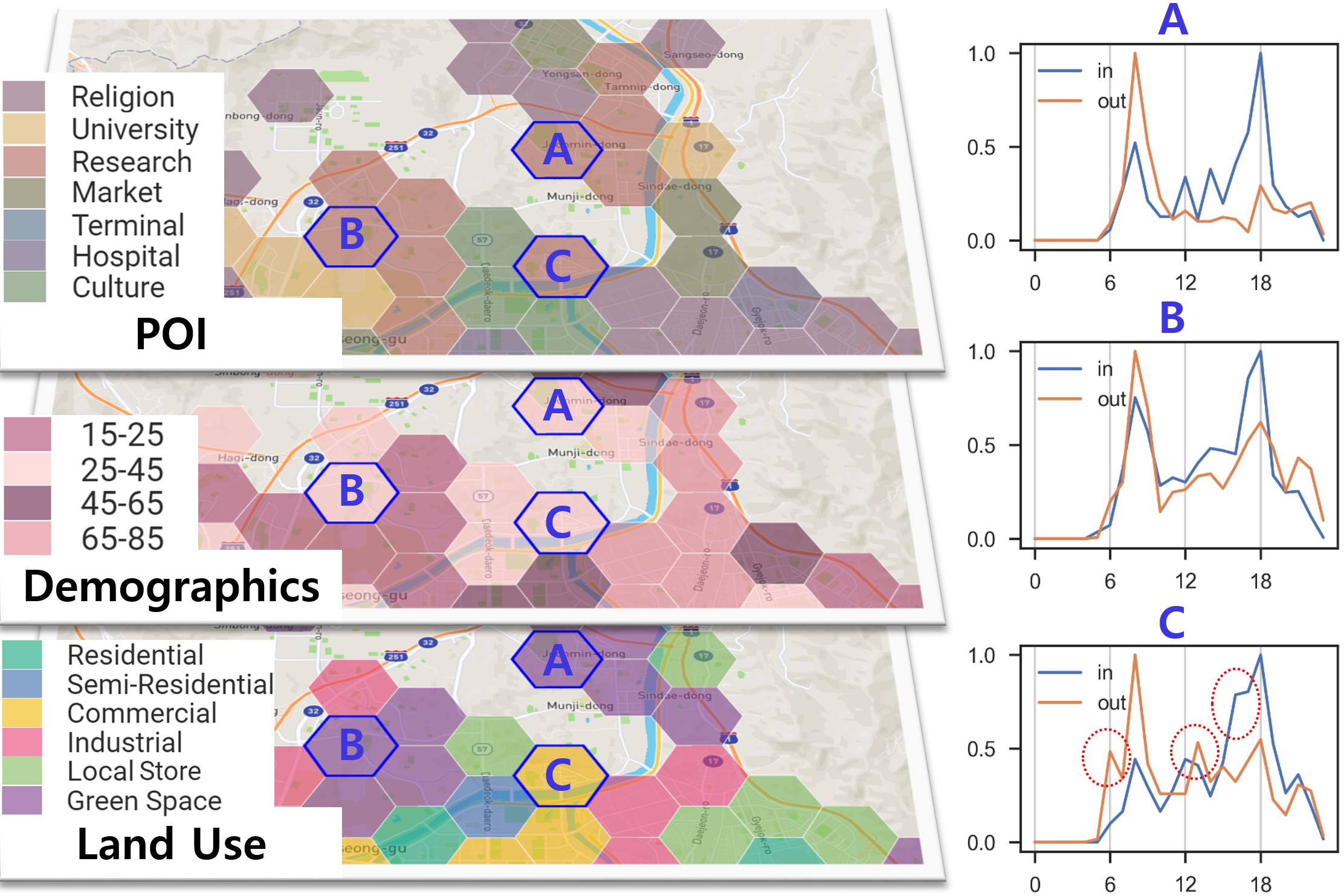}
    \caption{Similar travel patterns of areas with the same POI and demographics and another data layer of land use (color represents the majority group).}
    \label{fig:intro}
\vspace{-5mm}
\end{figure}

A reliable short-term transportation demand prediction supports the authorities in improving the capability of transportation systems.
However, this task is challenging because the transit demand prediction depends on the complex spatial and temporal correlation~\cite{morency2007measuring}. 
With the development of deep learning approaches to short-term transportation demand prediction, several models have been proposed to deal with the spatio-temporal complexity in transportation demand~\cite{luca2021survey,shi2015convolutional,zhang2017deep,lin2019deepstn+,liu2021community}.
For better prediction, first,  temporal dependency should be considered  
as each area shows similar transportation demand according to similar time zones on the same day of the week, as well as historical transportation demand for each timestep~\cite{foell2016regularity,ma2013mining}. 
Secondly, spatial dependency should also be incorporated.
According to Tobler's first law of geography~\cite{tobler1970computer}, the travel demand of an area has significant spatial relation to the adjacent areas.
Moreover, even distant areas with a similar urban environment (e.g., land use, point-of-interests (POIs)) show similar travel patterns due to the comparable purpose of visit~\cite{chen2009diurnal,gan2020understanding,ling2022sthan}. 
\looseness=-1


A handful of deep learning models incorporate areal features while learning spatio-temporal correlation for transportation demand prediction, such as POI distribution~\cite{lin2019deepstn+,pan2019urban,geng2019spatiotemporal,tang2021multi} and demographics~\cite{liu2021community}. 
Despite promising results, we argue that one important aspect has been overlooked in those efforts; 
they represented an area with  a very few features. 
Urban characteristics are polymorphic~\cite{harrison1996re}, and an area should be defined considering physical, social, and cultural aspects through multiple data layers such as land use, sociodemographics, building types, POI distribution, and transportation infrastructures~\cite{taylor2009nature,ewing2010travel}. 
For example, three areas (A, B, C) in Fig.1 show a similar travel pattern (more alighting in the morning and more boarding in the evening). 
This is expected as they all have many offices, and people aged 25-45 are commuting to those areas during workdays. 
However, there exists a difference observed in the area \textbf{C} -- small peaks in the early morning (6h) and daytime (13h, 16h), which could be driven by other areal urban features such as land use. 
Thus, to better predict future travel demand, 
it is necessary to devise a new method that can leverage multiple areal features by considering their significance and correlation. 


In this paper, we propose a spatio-temporal multi-feature-aware graph convolutional recurrent network (ST-MFGCRN) model that ensembles multiple areal similarity knowledge while preventing overfitting by sentinel attention\footnote{{The code, data, and supplemental materials are available in this anonymous repository:} \url{https://anonymous.4open.science/r/ST-MFGCRN-251B/}}. 
First, to capture the spatial relatedness of different time intervals, we split time-series input data into three kinds of time units (closeness, period, and trend).
Our model ingests each time-unit input with spatio-temporal embedding to capture trivial spatio-temporal patterns.
Then, we use our multi-feature graph convolutional recurrent neural network to understand multiple areal similarity graphs while finding spatio-temporal correlations.
Here, the areal similarity is computed by an attention-based mechanism from areal features.
This attention mechanism is improved with sentinel variables that allow each area to take partial attention when the areal feature is not helpful.
Finally, we merge the three time-unit outputs by our proposed weighted fusion mechanism to mix them properly. 
We evaluate the ST-MFGCRN model with other existing schemes against two real-world datasets.
Here we summarize our contributions:
\begin{itemize}
    \item {We propose a novel deep-learning architecture for short-term transportation demand prediction that incorporates multiple areal features while learning spatio-temporal correlation.}
    \item We analyze the impact of areal features to extract the best feature combination and highlight which areal features drive the best performance.
    \item Evaluation results show that the proposed ST-MFGCRN outperforms state-of-the-art base-lines such as DeepSTN+ and GMAN up to 7-8\%, and is also resilient when many areal features are computed together.
\end{itemize}

\section{Related Work}




\subsection{Spatio-temporal prediction methodology}

\textbf{For Euclidian dataset}, a convolutional neural network (CNN) based spatio-temporal model is mainly used~\cite{luca2021survey}.
ConvLSTM~\cite{shi2015convolutional} is a model modified for time series data by making 2D convolution in the matrix multiplication part of the weight parameter of a LSTM.
STResNet~\cite{zhang2017deep} and DeepSTN+~\cite{lin2019deepstn+} learn spatial information with 2D CNN for closeness, period, and trend, respectively, and time series temporal information with 1D CNN.
DeepSTN+ is a slightly advanced form compared to STResNet in that it leverages a ConvPlus network, but both are similar as they are CNN-based models.

\noindent\textbf{For non-Euclidian dataset}, a graph convolutional network (GCN) based spatio-temporal model is mainly used ~\cite{jiang2021graph,tang2021multi}. 
DCRNN~\cite{li2017diffusion} proposes an RNN model through random walk-based diffusion convolution based on the distance of traffic sensor to learn graph-based spatio-temporal information.
ASTGCN~\cite{guo2019attention} leverages spatial and temporal attentional graph convolution mechanism on closeness, period, and trend components for road traffic prediction.
GMAN~\cite{zheng2020gman} creates performs spatial attention with node embedding, and temporal attention, and combine with gated fusion. 

\subsection{Transit prediction with extra knowledge}\label{sec:RW1}

A handful of researchers~\cite{lin2019deepstn+,zhang2017deep,pan2019urban,liu2021community,tang2021multi,yuan2021effective}
have attempted to increase the prediction accuracy by using ancillary information about the area, such as POI or demographics.
There exist two major methods to utilize areal features---concatenating each areal feature along with input data or generating a similarity graph between areas and applying graph convolution.
The former is a method of obtaining a specific pattern from the transit demand value itself, and the latter differs in that it utilizes information from areas with similar characteristics to inference when performing prediction for each area.
\cite{lin2019deepstn+,zhang2017deep,pan2019urban} correspond to the former, although there are modifications such as using more fully connected layers or conducting addition instead of concatenation.
\cite{liu2021community,tang2021multi,yuan2021effective,geng2019spatiotemporal} utilize Pearson correlation, cosine similarity, or community detection by the Louvain algorithm to construct similarity matrices.


\subsection{Multi-graph spatio-temporal methodology}
There are a few approaches that merges multiple graphs for spatio-temporal computation. 
DMVST-Net\cite{yao2018deep} suggests semantic view on top of spatial view and temporal view by creating data-driven semantic similarity graph by Dynamic Time Wraping for taxi demand prediction.
ST-MGCN\cite{geng2019spatiotemporal} proposes multi-graph convolution network consists of neighborhood, functional similarity, and spatial connectivity for ride-hailing demand forecasting.
Wang et al.\cite{wang2021forecasting} proposes heterogeneous multi-graph convolution network consists of spatial adjacency, transport connectivity, and contextual similarity for ambulance demand forecasting.
AGCAN\cite{li2021adaptive} leverages data-driven attention-based adaptive graph to supplement along with physical connectivity graph for road traffic prediction.


\section{Problem formulation}

We define that the number of transportation areas is $N$, the number of input channels is $C_x$ (e.g. $C_x=2$ for in/out demand of each area), and the demand (e.g., number of passengers who aboard/depart the bus) on each timestamp at $X_t \in \mathbb{R}^{N \times C_x}$, and the spatial proximity graph between areas is $\mathcal{G}^P$. 
Assume we want to use $K_N$ types of areal features, where $k$-th feature is $F_k \in \mathbb{R}^{N \times V_k}$, where $V_k$ is the number of components of $F_k$ (e.g. $V_{LU}=3$ when the land use feature $F_{LU}$ consists of commercial/residential/office).
The transportation demand prediction problem can be defined as learning a model $f_{\text{model}}$ that predicts the next-step value from historical values as: 
\begin{equation}
f_{\text{model}}(X_{1}, X_{2}, ..., X_{t-1}; \mathcal{G}^P, F_{\{1,...,K_N\}}) \rightarrow X_{t}
\end{equation}

\section{Method}


        



\subsection{Overview}

\begin{figure*}[h]
\vspace{-5mm}
\centering
\subfloat[ST-MFGCRN architecture\label{fig:model}]{\includegraphics[width=0.4\textwidth]{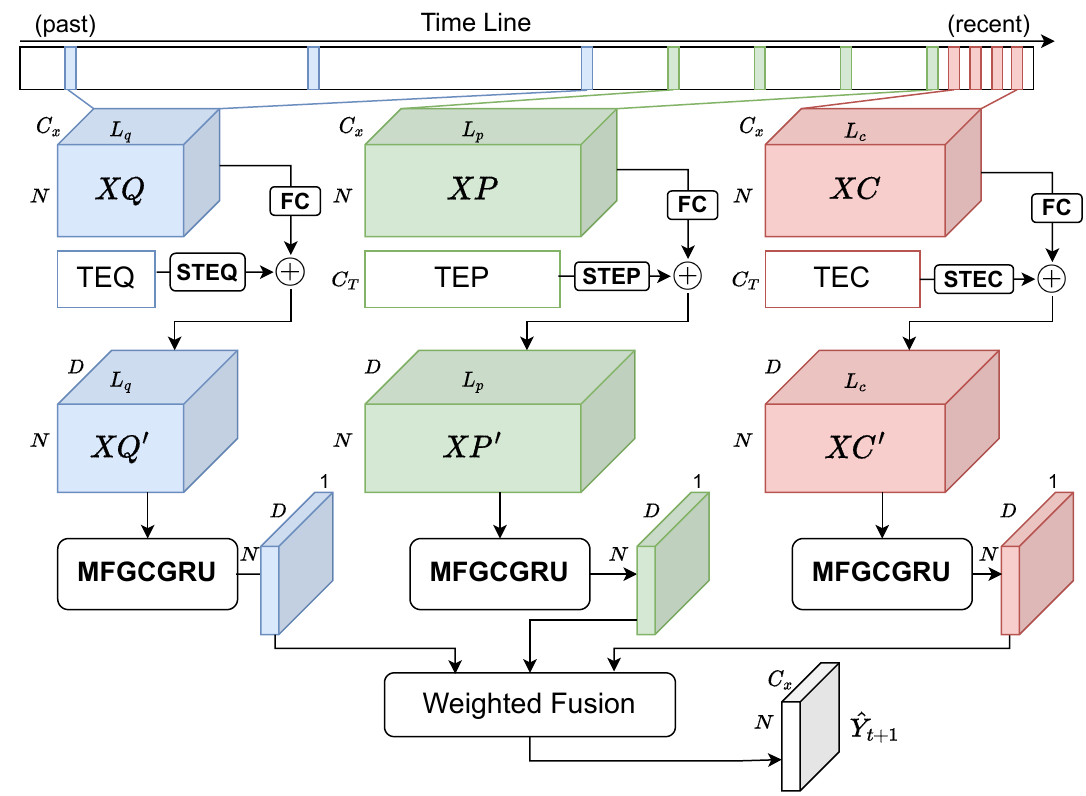}} 
\subfloat[Multi-Feature aware GCGRU (MFGCGRU)\label{fig:mfgcgru}]{\includegraphics[width=0.6\textwidth]{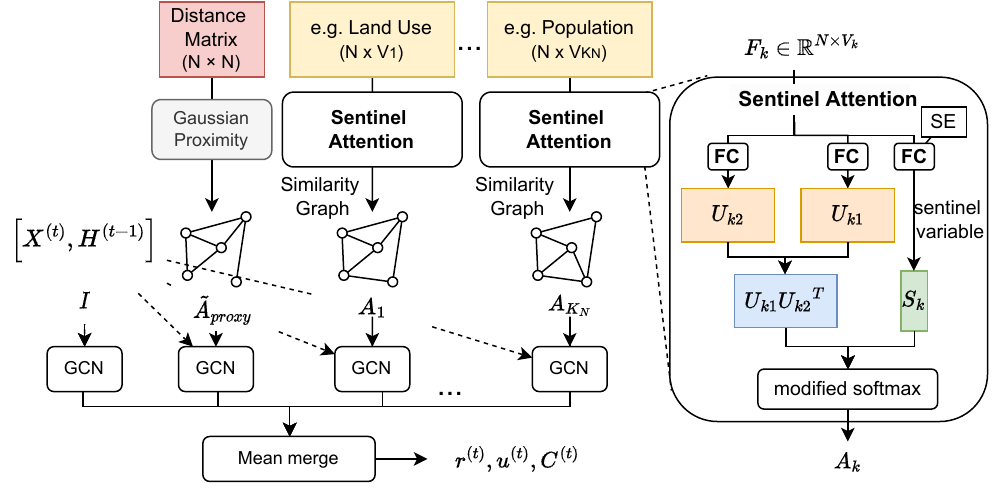}}
\caption{(a) Proposed ST-MFGCRN architecture and (b) MFGCGRU cell. TEQ, TEP, TEC represents the temporal embedding (TE) for trend, period, and closeness modules which convert into spatio-temporal embedding (STE).
} \label{fig:proposed}
\vspace{-2mm}
\end{figure*}

We propose spatio-temporal multi-feature graph convolutional recurrent network (ST-MFGCRN) described in Fig.~\ref{fig:model}.
Our method splits the historical value into three different time units -- closeness (recent minutes), period (daily), and trend (weekly) similar to previous approaches\cite{lin2019deepstn+,zhang2017deep,zhang2016dnn} as follows: 
\begin{equation}
\begin{aligned}
    &\boldsymbol{XC} = \{ X_{t}, ..., X_{t-L_c+1} \} \in \mathbb{R}^{N \times L_c \times C_x}, \\
    &\boldsymbol{XP} = \{ X_{t-T_p+1}, ..., X_{t-T_p L_p +1} \}  \in \mathbb{R}^{N \times L_p \times C_x}, \\
    &\boldsymbol{XQ} = \{ X_{t-T_q+1}, ..., X_{t-T_q L_q +1} \}  \in \mathbb{R}^{N \times L_q \times C_x}
\end{aligned}\label{eq:CPT}
\end{equation}

where $L_c$, $L_p$, and $L_q$ are the number of timesteps of closeness, period, and trend values, respectively. $T_c$ is the number of close sequences, and $T_p$ is the number of steps between periods while $T_q$ is that for trends. 
For hidden input and output between all modules, we use $D$ as the embedding dimension in common.


\subsection{Spatio temporal embedding (STE)}
Spatio temporal embedding (STE) serves to capture and correct trivial spatio-temporal bias for each area.
Compared to GMAN~\cite{zheng2020gman} that leverages STE that serves similar to positional encoding, our model directly adds to embedded input value.
We preprocess the temporal embedding (TE) as $\text{TE}_t \in \mathbb{R}^{C_T}$ by concatenating each one-hot embedding of a time unit, where $C_T=7+24+4+1$: weekday (7), hour-of-day (24), 15-min-unit-of-hour (4), national holiday (1). 
Then, we apply a two-stacked fully connected layer $f_{\text{TE}}: \mathbb{R}^{C_T} \rightarrow \mathbb{R}^D$ to convert TE into $D$-dimensional hidden embedding. 
Moreover, we define a trainable spatial embedding $\text{SE} \in \mathbb{R}^{N \times D}$ and make the spatio-temporal embedding as  $\text{STE}_t = \text{SE} + f_{\text{TE}}(\text{TE}_t) \in \mathbb{R}^{N \times D}$. 
Meanwhile, we extend the initial traffic input $X_t \in \mathbb{R}^{N \times C_x}$ into $f_{\text{in}}(X_t) \in \mathbb{R}^{N \times D}$ by applying a two-stacked fully connected layer $f_{\text{in}} : \mathbb{R}^{C_x} \rightarrow \mathbb{R}^D$, and produce the traffic value embedding normalized with STE as $X'_t = f_{\text{in}}(X_t) + \text{STE}_t \in \mathbb{R}^{N \times D}$ and use it as a next input of a module. Here, we share the same SE and $f_{\text{TE}}$ for different time units, while $f_{\text{in}}^{(c)},f_{\text{in}}^{(p)},f_{\text{in}}^{(q)}$ differently used for each $\boldsymbol{XC}, \boldsymbol{XP}, \boldsymbol{XQ}$.

\subsection{Multi-feature aware GCGRU (MFGCGRU)}
\label{proxy}


We propose a multi-feature aware GCGRU (MFGCGRU) by extending graph convolutional gated recurrent unit to compute multiple areal features inside GRU cell.
At first, the model computes spatial proximity $\mathcal{G}^P$(${A}_{proxy}$) by a graph convolution.
A proximity matrix is calculated from distances witsh a Gaussian filter as ${A}_{proxy}[i,j] = \text{exp}{(- (d_{ij}/\sigma)^2)}$, where the $d_{ij}$ is the distance between $i$-th and $j$-th areas and $\sigma$ is the standard deviation of distance values. 
After we conduct row-normalization as $\tilde{A}_{proxy}$, we use this matrix as one of graphs computed in MFGCGRU. 



Next, in order to extract areal similarity matrix $A_k \in \mathbb{R}^{N \times N}$ from areal feature $F_k$ to be used in MFGCGRU, we leverage attention mechanism as follows:

\begin{equation}
\begin{aligned}
U_{k1} = \text{ReLU}(\boldsymbol{W}_{k1} F_k) \in \mathbb{R}^{N \times D}, \quad
U_{k2} = \text{ReLU}(\boldsymbol{W}_{k2} F_k) \in \mathbb{R}^{N \times D}, \\
e_k = {U_{k1} {U_{k2}}^T } / \sqrt{D} \in \mathbb{R}^{N \times N}, \qquad
A_k[i, j] = \frac{\text{exp}(e_k[i,j])}{\sum_o{\text{exp}(e_k[i,o])}}
\end{aligned}
\end{equation}

\noindent, where $\boldsymbol{W}_{k1}, \boldsymbol{W}_{k2} \in \mathbb{R}^{D \times V_k}$ are trainable parameters for each feature type $k$. 
$e_k$ is basically represents the similarity between $U_{k1}$ and $U_{k2}$ by applying dot product for each $N \times N$ area pairs.
%

However, as more areal features are added, the model is congested and do not reflect the areal similarity due to the curse of dimensionality\cite{indyk1998approximate}.
Inspired by a sentinel attention~\cite{lu2017knowing,park2020st}, we calculate the attention score for each area can be less than 1 by applying sentinel variable $S_k$, which allows an area to take partial attention when the areal knowledge is not beneficial. Our proposed sentinel attention is defined as follows:
\begin{equation}
\begin{aligned}\label{eq:sent}
S_{k} &= f_{\text{sent}}(F_k \mathbin\Vert \text{SE} ) \in \mathbb{R}^{N}, \\
A_{k}[i, j] &= \frac{\text{exp}(e_k[i,j])}{S_{k}[i] + \sum_o{\text{exp}(e_k[i,o])}}
\end{aligned}
\end{equation}

\noindent, where $f_{\text{sent}}$ is a two-stacked fully connected layer with ReLU activation (which keeps $S_k \geq 0$). We apply concatenation ($F_k \mathbin\Vert \text{SE}$) to calculate sentinel variable differently for each area using spatial embedding.

\begin{equation} \label{eq:mgcrn}
\begin{aligned}
\boldsymbol{r}^{(t)}&=\quad \sigma\left(\frac{1}{K}\sum_{k=1}^{K}{ A_{k} \left(\boldsymbol{X}^{(t)} \mathbin\Vert \boldsymbol{H}^{(t-1)}\right) \boldsymbol{W}_{rk}}  +\boldsymbol{b}_{r} \right) \quad \\
\boldsymbol{u}^{(t)}&=\sigma\left(\frac{1}{K}\sum_{k=1}^{K} {A_{k} \left(\boldsymbol{X}^{(t)} \mathbin\Vert \boldsymbol{H}^{(t-1)}\right) \boldsymbol{W}_{uk}} +\boldsymbol{b}_{u}\right)\\
\boldsymbol{C}^{(t)}&=\tanh \left(\frac{1}{K}\sum_{k=1}^{K} {A_{k} \left(\boldsymbol{X}^{(t)} \mathbin\Vert \left(\boldsymbol{r}^{(t)} \odot \boldsymbol{H}^{(t-1)}\right)\right) \boldsymbol{W}_{Ck}} + \boldsymbol{b}_{c} \right) \quad\\
\boldsymbol{H}^{(t)}&=\boldsymbol{u}^{(t)} \odot \boldsymbol{H}^{(t-1)}+\left(1-\boldsymbol{u}^{(t)}\right) \odot \boldsymbol{C}^{(t)}.
\end{aligned}
\end{equation}

Finally, we ensemble multiple graph convolution inside GRU cell by applying mean merge of multiple graphs as Equation~\ref{eq:mgcrn}, where $\boldsymbol{W}_{rk},\boldsymbol{W}_{uk},\boldsymbol{W}_{Ck} \in \mathbb{R}^{(D_x+D_h)\times D_h}$ (for $k=1,...,K$) and $\boldsymbol{b}_{r},\boldsymbol{b}_{u},\boldsymbol{b}_{C} \in \mathbb{R}^{ D_h}$ are trainable parameters, and $\odot$ is Hadamard product. 
We also apply an identity matrix ($I$) on top of a proximity matrix and $K_N$ area similarity matrices as a residual computation, thus $K=2+K_N$ as a default.

\subsection{Weighted fusion network}

From each MFGCGRU for $\boldsymbol{XC}$, $\boldsymbol{XP}$, $\boldsymbol{XQ}$, we get an output of $H_C, H_P, H_Q \in \mathbb{R}^{N\times D}$, respectively. 
We apply a fully connected layer and a softmax to calculate the weights of importance, and produce final output with a two-stacked fully connected layer $f_{\text{fusion}}: \mathbb{R}^D \rightarrow \mathbb{R}^{C_x}$ as follows, where $\boldsymbol{W}_C, \boldsymbol{W}_P, \boldsymbol{W}_Q \in \mathbb{R}^{1 \times D}$ are trainables:

\begin{equation}
\begin{aligned}
e_m = \boldsymbol{W}_{m} H_m \in \mathbb{R}^N ( m \in \{C,P,Q\}), \quad
\alpha_{m} = \frac{\text{exp}(e_m)}{\sum_j^{\{C,P,Q\}}{\text{exp}(e_j)}} \in \mathbb{R}^N, \\
\hat{Y}_{t+1} = f_{\text{fusion}}\left(   \sum_j^{\{C,P,Q\}}{\alpha_j \odot H_j}  \right) \in \mathbb{R}^{N \times {C_x}}.
\end{aligned}
\end{equation}

\subsection{Objective function}

The objective function is defined as 
 $\mathcal{L}(\theta) =  \frac{1}{N} \sum_i^N {| Y^i_{t+1} - \hat{Y}^i_{t+1}|}$ where $Y_t \in \mathbb{R}^{N \times C_x}$, which is the \textit{mean absolute error} (L1) loss.


\section{Evaluation settings}

\subsection{Dataset}

We use two datasets in our experiments: our constructed BusDJ dataset and a benchmark TaxiBJ dataset. 
For the BusDJ dataset, 
we use the bus transit data of boarding and alighting demand in a 15-minute unit which we obtain from smart card data of Daejeon Metropolitan City. 
We construct the bus transit dataset of $N=102$ hexagonal 800m grid areas except the areas containing no bus stop (e.g., hills, rivers).
Then, we leverage six areal features representing land use, transit, and demographic features: (1) land use type (LU, $V_{LU} = 6$), (2) place-of-interest (POI, $V_{POI} = 29$), (3), bus transit infrastructure (BUS, $V_{BUS} = 2$), (4) population (POP, $V_{POP} = 4$), (5) enterprise and employments of each business type (ENT, $V_{ENT} = 6$), and (6) building area (BD, $V_{BD} = 7$). The areal features are provided by the open data service of government institutions~\cite{LocalData,KoreaNSDI,SGIS}. 
For benchmark TaxiBJ \cite{pan2019urban} dataset, 
we use two available areal features provided in the original dataset.
We extract the 20 most frequent POI features among the original 648 types of POIs ($V_{POI} = 20$). 
We also use two ROAD features (the number of roads and lanes) as original data ($V_{ROAD} = 2$).
The detailed description of both data is listed in Table~\ref{tab:dataset}.


\begin{table}[h]
\centering
\caption{Details of the datasets.\label{tab:dataset}}
\begin{tabular}{l  c c}
\toprule
\textbf{Dataset} & \textbf{BusDJ} & \textbf{TaxiBJ} \\ \midrule
$N$ (area size) & 102 (800m Hexa) & 146 (1km Square)\\ 
Train timespan & 3/1/2019 - 4/30/2019 & 2/1/2015 - 6/2/2015 \\ 
Val timespan & 5/1/2019 - 5/31/2019 & 6/3/2015 - 6/16/2015 \\ 
Test timespan & 6/1/2019 - 6/30/2019 & 6/17/2015 - 6/30/2015 \\ 
Time interval & 15min & 30min \\ 
\# timestamps & 6832 & 7152 \\ 
Average demand (in/out) & 32.05/31.41 & 626.10/625.35 \\ 
Target prediction time & 8h - 21h (14h)  & 0h - 23h (24h) \\ 
\# areal features & 6 & 2 \\ \bottomrule
\end{tabular}
\vspace{-5mm}
\end{table}


\subsection{Experiment settings}

We set the number of time sequences as $L_c = 6, L_p = 7, L_q = 3$ in Eq.~\ref{eq:CPT}, and $C_x = 2$ for boarding and alighting demand.
In training, we conduct min-max normalization on demand and areal features.
We set the hidden embedded dimension for our model as $D=64$. For other baselines, we use their default settings.
We conduct our experiments on a single NVIDIA TITAN RTX 24GB environment with Tensorflow v1.15.1.
For each model on each setting, we test five times and record the mean of each result.
We adopt an ADAM optimizer with a learning rate of 0.01 and a batch size of 32. We apply early stopping with validation loss with 15 patience epochs.
We evaluate models with \textit{root mean squared error} (RMSE) and \textit{mean absolute error} (MAE).
We measure the error of $C_x$ input channels altogether.

\subsection{Baselines} 

We compare the proposed model (ST-MFGCRN) with various methods for public transit prediction.
We first use data-based prediction methods without model training as basic baselines: Trend Mean, Period Mean, Closeness Mean, and Last Repeat.
We set CNN-based time-series deep learning models, such as ConvLSTM~\cite{shi2015convolutional}, STResNet~\cite{zhang2017deep}, and DeepSTN+~\cite{lin2019deepstn+}. 
We also compare our model with the graph-based deep learning models such as DCRNN~\cite{li2017diffusion}, ASTGCN~\cite{guo2019attention}, 
STMGCN~\cite{geng2019spatiotemporal}, 
and GMAN~\cite{zheng2020gman} by using proximity information $A_{proxy}$ described in Section~\ref{proxy}. 
For STMGCN, we use cosine-similarity based adjacency matrix for each feature for graph computation.
For ASTGCN and STMGCN, we use closeness, period, trend component same as our model. 
For other baselines, we use the default settings.

\section{Results}

We first analyze the feature influence by adding or removing each areal feature from our model to extract the most important areal feature combination. 
Then, we evaluate the performance of ST-MFGCRN by comparing it with baseline models on the BusDJ and TaxiBJ datasets. 

\subsection{Analysis of feature significance} \label{sec:feature}

\begin{table}[h]
\centering
\noindent
\begin{minipage}[c]{0.6\textwidth}
\centering

                \captionof{table} {Feature influence on BusDJ (RMSE). Results of important features  are underlined. } \label{tab:f1}
                
                \begin{tabular}{l|l|c|c}
                \toprule
                  &  Feature add/remove & $-\mathcal{G}^P$ & $+\mathcal{G}^P$ \\ \midrule
                $K_N=0$ & N/A &  & 8.5032 \\ \midrule
                $K_N=1$ & +LU & 8.5000 & \underline{8.4625} \\ 
                 & +POI & 8.5409 & \underline{8.4675} \\ 
                 & +BUS & 8.5141 & 8.4747 \\ 
                 & +POP & 8.5063 & 8.5119 \\ 
                 & +ENT & 8.5100 & \underline{8.4638} \\ 
                 & +BD & 8.5290 & 8.5031 \\ \midrule
                $K_N=5$ & (all)--LU & 8.4941 & 8.4635 \\ 
                 & (all)--POI & 8.4939 & \underline{8.4784} \\ 
                 & (all)--BUS & 8.4976 & 8.4670 \\ 
                 & (all)--POP & 8.4976 & 8.4619 \\ 
                 & (all)--ENT & 8.4937 & \underline{8.4816} \\ 
                 & (all)--BD & 8.4782 & \underline{8.4963} \\ \midrule
                $K_N=6$ & (all) &  8.5005 & 8.4606 \\ \midrule
                \textbf{Best} & +BD, +ENT, +POI, +LU & & \textbf{8.4587} \\ 
                \bottomrule
                \end{tabular}

\end{minipage}
\hspace{3mm}
\begin{minipage}[c]{0.35\textwidth}
\centering
                \captionof{table} {Feature influence on TaxiBJ dataset.} \label{tab:f2}
                \begin{tabular}{c|c|c|c}
                \toprule
                $\mathcal{G}^P$ & $F_{POI}$ & $F_{ROAD}$ & RMSE    \\ \midrule
                --         & --   & --    & 99.849 \\ 
                +         & --   & --    & 95.487 \\ 
                +         & +   & --    & 93.860 \\ 
                +         & --   & +    & 92.782 \\ 
                +         & +   & +    & \textbf{92.767} \\ \bottomrule
                \end{tabular}
\end{minipage}
\vspace{-2mm}
\end{table}

To examine to what extent individual features improve prediction performance, we conduct a feature influence study by adding or removing features.

For the BusDJ dataset, Table~\ref{tab:f1} shows a significant error decrease when the proximity graph $\mathcal{G}^P$ is provided, implying that geographical closeness essentially captures the basic spatial correlation and improves the performance.
The experiment results by all combinations (when $K_N = 2,3,4,5$) are omitted due to the space limit. 
We find that LU, ENT, and POI are the most significant areal features when added, as their errors decrease the most from \{$K_N=0,+\mathcal{G}^P$\} model. 
On the other hand, when we evaluate the most significant areal features when removed, 
we observe BD, ENT, and POI increase the most errors from \{$K_N=6,+\mathcal{G}^P$\} model.
%
Throughout all experiments, the best setting is found as a combination of BD, ENT, POI, and LU.
%
%
%




For TaxiBJ dataset, Table~\ref{tab:f2} shows the POI and ROAD features do improve the performance significantly. The model shows the best result when it takes all the areal features as well as the proximity graph.

These results imply that each areal feature provides unique information about the urban environment, and multiple views of urban area supplements the overall performance.

\subsection{Performance comparison}

\begin{table}[h]
\caption {Performance comparison.}
\label{tab:comparison} 
\centering
\begin{tabular}{l|rr|rr}
\toprule    & \multicolumn{2}{c|}{BusDJ}   & \multicolumn{2}{c}{TaxiBJ} \\ \hline
\multicolumn{1}{c|}{Model} & \multicolumn{1}{c}{RMSE} & \multicolumn{1}{c|}{MAE} & \multicolumn{1}{c}{RMSE} & \multicolumn{1}{c}{MAE}    \\ \midrule
        Trend Mean & 11.3647 & 6.8445 & 307.430 & 199.915 \\ 
        Period Mean & 15.3843 & 8.8179 & 309.706 & 198.528 \\ 
        Closeness Mean & 15.8035 & 8.9490 & 225.092 & 159.634 \\ 
        Last Repeat & 13.6378 & 8.6482 & 116.877 & 72.842 \\ \midrule
        ConvLSTM & 9.2843 & 5.9005 & 100.939 & 57.500 \\ 
        STResNet & 9.2875 & 5.9241 & 103.689 & 63.305 \\ 
        DeepSTN+ & 9.2215 & 5.8746 & 107.319 & 64.830 \\ \midrule
        DCRNN & 9.5296 & 6.1572 & 107.138 & 66.566 \\ 
        ASTGCN & 10.7053 & 6.5875 & 126.197 & 75.555 \\ 
        STMGCN & 10.0883 & 6.3992 & 110.993 & 71.346\\
        GMAN & 9.1381 & 5.8662 & 102.276 & 56.109 \\ \midrule
        DCRNN+STE+CPT & 8.7775 & 5.6461 & 106.756 & 62.876 \\ 
        GMAN+STE+CPT & 8.9188 & 5.7136 & 113.828 & 67.689 \\ \midrule
        \textbf{ST-MFGCRN (best*)} & \textbf{8.4635} & \textbf{5.4987} & \textbf{92.767} & \textbf{52.443} \\
        \bottomrule

        \multicolumn{5}{r}{\textbf{*}BusDJ: BD, ENT, POI, LU, TaxiBJ: POI, ROAD}
        

\end{tabular}
\vspace{-5mm}
\end{table}


Table~\ref{tab:comparison} shows the performance comparison of models on BusDJ and TaxiBJ datasets.
Our proposed ST-MFGCRN shows an improved performance of 7\% on BusDJ dataset and 8\% on the TaxiBJ dataset compared to the state-of-the-art baselines such as DeepSTN+ and GMAN in terms of RMSE.
Among the basic baselines, Trend Mean shows the best performance, which implies that the travel patterns are similar for the same hour and weekday every week.
On the other hand, Closeness Mean and Last Repeat show high errors, indicating the need for a temporal correlation training model.
Among the Euclidian area-based models, DeepSTN+ shows the highest performance as DeepSTN+ is more advanced than STResnet or ConvLSTM.
In addition, ConvLSTM shows better performance than STResNet which leverages LSTM instead of 1D-CNN to find the temporal correlation.
Among the non-Euclidian models, GMAN shows the highest performance compared to DCRNN and ASTGCN as it leverages spatial attention which allows attention on all other areas rather than graph convolution with fixed proximity values.

We also conduct experiments when we expand DCRNN with STE and closeness/period/trend (CPT) unit and GMAN with CPT unit, and find each component improves performance on these baselines.
However, our ST-MFGCRN still outperforms as it leverages multiple features with sentinel attention  and applies weighted fusion to utilize these components.

\section{Conclusion}

We propose ST-MFGCRN which allows exploiting various areal features during spatio-temporal training to tackle transportation demand prediction problems.
To capture correlations depending on areal characteristics and spatial adjacency, ST-MFGCRN incorporates multiple similarity matrices calculated from different areal features and a distance-based proximity matrix.
To calculate the similarity matrix of each areal feature, we leverage our proposed sentinel attention, which plays a role in taking partial attention when the knowledge is not helpful and harmoniously mixes various features.
ST-MFGCRN captures temporal correlation by different time unit inputs of closeness, period, and trend, and spatio-temporal dependencies are also captured by spatio-temporal embedding. 
Our model outperforms the state-of-the-art baselines on the real-world dataset that we construct as well as on the benchmark dataset while successfully leveraging multiple areal features.


\bibliographystyle{ieeetr}

\bibliography{my-base,multigraph}



\end{document}